\documentclass[final]{cvpr}

\usepackage{times}
\usepackage{epsfig}
\usepackage{graphicx}
\usepackage{amsmath}
\usepackage{amssymb}
\usepackage{comment}
\usepackage[inline]{enumitem}
\usepackage[caption=false]{subfig}
\usepackage{algorithmic}
\usepackage{algorithm}

\usepackage[pagebackref=true,breaklinks=true,colorlinks,bookmarks=false]{hyperref}
 \usepackage{array,multirow,graphicx}

\def\cvprPaperID{6241} % *** Enter the CVPR Paper ID here
\def\confYear{CVPR 2021}

\newcommand{\cL}{\mathcal{L}}

\newcommand{\cN}{\mathcal{N}}

\newcommand{\cO}{\mathcal{O}}

\newcommand{\cC}{\mathcal{C}}

\newcommand{\bh}{\mathbf{h}}

\newcommand{\bo}{\mathbf{o}}
\newcommand{\bl}{\mathbf{l}}

\begin{document}

%%%%%%%%% TITLE
\title{\textit{DyGLIP}: A Dynamic Graph Model with Link Prediction for \\ 
Accurate Multi-Camera Multiple Object Tracking}

\author{Kha Gia Quach$^{1}$, Pha Nguyen$^{2}$, Huu Le$^{3}$, Thanh-Dat Truong$^{4}$, Chi Nhan Duong$^{1}$ \\ Minh-Triet Tran$^{5}$, Khoa Luu$^{4}$ \\
 $^{1}$ Concordia University, CANADA \quad 
 $^{2}$ VinAI Research, VIETNAM \\
 $^{3}$ Chalmers University of Technology, SWEDEN \quad
 $^{4}$ University of Arkansas, USA \\
 $^{5}$ University of Science, VNU-HCM, VIETNAM \\
\tt\small$^{1}$\{dcnhan, kquach\}@ieee.org, $^{2}$ v.phana@vinai.io, $^{3}$ huul@chalmers.se, \\ \tt\small  $^{4}$\{tt032, khoaluu\}@uark.edu,  $^{5}$tmtriet@fit.hcmus.edu.vn
}

\maketitle

\newcommand{\cV}{\mathcal{V}}
\newcommand{\cE}{\mathcal{E}}
\newcommand{\cB}{\mathcal{B}}
\newcommand{\cG}{\mathcal{G}}

%%%%%%%%% ABSTRACT
\begin{abstract}
Multi-Camera Multiple Object Tracking (MC-MOT) is a significant computer vision problem due to its emerging applicability in several real-world applications. Despite a large number of existing works, solving the data association problem in any MC-MOT pipeline is arguably one of the most challenging tasks. Developing a robust MC-MOT system, however, is still highly challenging due to many practical issues such as inconsistent lighting conditions, varying object movement patterns, or the trajectory occlusions of the objects between the cameras. To address these problems, this work, therefore, proposes a new Dynamic Graph Model with Link Prediction (DyGLIP) approach
\footnote{Visit \url{https://github.com/uark-cviu/DyGLIP} for the implementation of DyGLIP.} to solve the data association task. Compared to existing methods, our new model offers several advantages, including better feature representations and the ability to recover from lost tracks during camera transitions. Moreover, our model works gracefully regardless of the overlapping ratios between the cameras. Experimental results show that we outperform existing MC-MOT algorithms by a large margin on several practical datasets. Notably, our model works favorably on online settings but can be extended to an incremental approach for large-scale datasets. 
\end{abstract}

%%%%%%%%% BODY TEXT
\section{Introduction}
\begin{figure}
    \centering
    \includegraphics[width = 0.93\columnwidth]{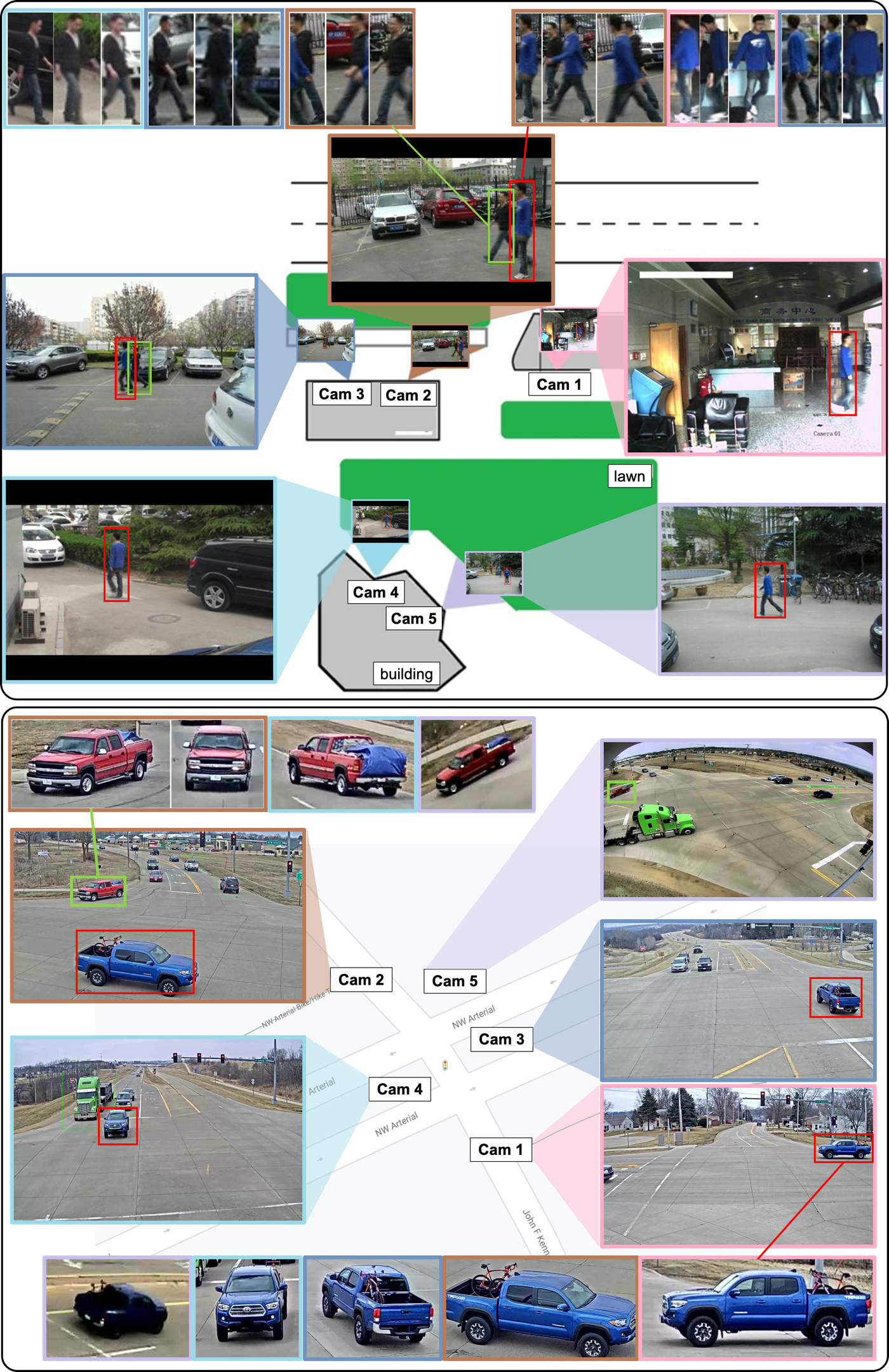}
    \caption{Top: Pedestrians in the MCT dataset \cite{chen2016equalized}, Bottom: Cars in the CityFlow dataset \cite{tang2019cityflow}.}
    \label{fig:abstractFig1}
    \vspace{-7mm}
\end{figure}
Multi-Camera Multiple Object Tracking (MC-MOT) plays an essential role in computer vision due to its potential in many real-world applications such as self-driving cars, crowd behavior analysis, anomaly detection, etc.
Although recent Multi-Camera Multiple Object Tracking (MC-MOT) methods have achieved promising results in several large-scale datasets, there are still many challenges that need to be addressed. Among them, data association, a crucial step in determining the performance of an MC-MOT pipeline, has attracted a considerable amount of attention lately. While the association itself is challenging in single-camera MOT applications, this task becomes even more difficult in MC-MOT settings due to possibly inconsistent lightning conditions or occlusion patterns between the cameras. For example, feature vectors, given by an off-the-shelf object detector and Re-Identification (Re-ID) models, associated with a particular object during its transition from an indoor to an outdoor environment can be totally different as shown in Fig. \ref{fig:abstractFig1}.
This breaks the connection between the trajectory representation of an object and its previous feature vectors, leading to wrong identity (ID) assignments. 
Failures in data association at a particular frame will potentially lead to long-term detrimental effects for an online tracking system. 
In practice, many circumstances that are much more complex than simply generate more new IDs which could significantly deteriorate the tracking accuracy. Therefore, improving data association plays a crucial role in determining the performance of an MC-MOT algorithm.

To tackle the problem of data association mentioned above, our work introduces a totally new perspective to tackle the association task in MC-MOT. Most previous works rely on feature vectors obtained from an underlying object detector and use them to solve the assignment problem, e.g., using nearest neighbors \cite{10.1145/3131885.3131921}, clustering \cite{8578730, Zhang2017MultiTargetMT} or solving an instance of non-negative matrix factorization \cite{he2020multi}. Instead, we propose to consider the problem of data association as \textit{a link prediction on a graph}, where the graph vertices are associated with the tracks. 
To construct our predictor, we introduce a new dynamic graph formulation that can take into account the temporal information of an object over a period of time and its relation to other objects. 
As shown in the experiments in Section \ref{ssec:ablation_study}, this dynamic graph formulation allows leveraging the feature representations and moving patterns of each object to improve ID assignment, leading to better performance compared to the state-of-the-art (SOTA) methods.

\vspace{2mm}
\noindent
\textbf{Contributions} The main contributions of our work can be summarized as follows. Firstly, a new MC-MOT framework is presented using the link prediction in conjunction with a dynamic graph formulation. We demonstrate that this new model significantly improves the association task in numerous MC-MOT datasets by a large margin. To the best of our knowledge, it is the first time link prediction and dynamic graph are used together in MC-MOT.
Secondly, the proposed dynamic graph will be incorporated with \textit{the attention mechanism}, allowing dynamically accumulating temporal and spatial information to result in a new graph embedding yielding highly accurate link prediction results.

%------------------------------------------------------------------------
\section{Related Work}

This section reviews some current methods on MC-MOT algorithms.
There is a large amount of research on single-camera MOT tackling the matching/association problem \cite{kim2015multiple, ren2018, Chen_2018_ECCV, maksai2019} or building an end-to-end framework that unified with the detector \cite{sun2019deep, xu2019deepmot, wang2019towards, bergmann2019tracking, chu2019famnet, zhou2020tracking, Li_2020_WACV, sun2020simultaneous, peng2020chained, xu2020train, yin2020unified}.
MC-MOT has recently received increasing attention intending to determine the global trajectory of all subjects in a multiple camera system simultaneously. Compared to single-camera MOT, MC-MOT is more challenging in the affinity stage. The difficulties may come from the significant changes in subject pose between cameras, a vast number of matching trajectories, or differences in object features. 

\textbf{Matching using spatio-temporal constraints:} Kumar et al. \cite{10.1145/3131885.3131921} use pre-defined spatio-temporal camera connections, represented by an adjacent moving time matrix, to search for the targeted person when he/she disappeared in a field of view. Similarly, Jiang et al. \cite{10.1145/3240508.3240663} estimate the camera topology to significantly reduced the number of matching candidates. Styles et al. \cite{9150757} propose several baseline methods that only solve a minor task that is classifying which camera will a disappeared target recur in. Jiang et al. \cite{10.1145/3240508.3240663} apply Gaussian Mixture Model to estimate the camera connectivity, therefore searching in probe tracklets set is less painful. Chen et al. \cite{7508485} formulate the assignment task as a min-cost flow matching problem in a network.

\textbf{3D matching on an overlapping field of view:} You and Jiang \cite{you2020realtime} introduce a method that works specifically on overlapping fields of view by estimating people's ground heatmap.  Chen et al. \cite{Chen_2020_CVPR} reconstruct 3D geometric information to overcome 2D matching's limitations efficiently.

\textbf{General matching approaches:} He et al. generate local tracklets offline in all single views to construct a similarity matrix and then estimate global targets and their trajectory by performing the Matrix Factorization method. Ristani and Tomasi \cite{8578730} adopt correlation clustering to solve the ID assignment task. Then two post-processing steps, i.e., interpolation and elimination, are also executed to fill gaps and filter indecision tracks. Yoon et al. \cite{8387029} not only revisit the Multiple Hypothesis Tracking algorithm, which maintains a set of track hypotheses all the time but also reduce its expensiveness by introducing a gating mechanism for tree pruning. Zhang et al. \cite{Zhang2017MultiTargetMT} cluster IDs by using the Re-Ranking algorithm \cite{8099872} on the global cost matrix.

\begin{figure*}
    \centering
    \includegraphics[width = 2\columnwidth]{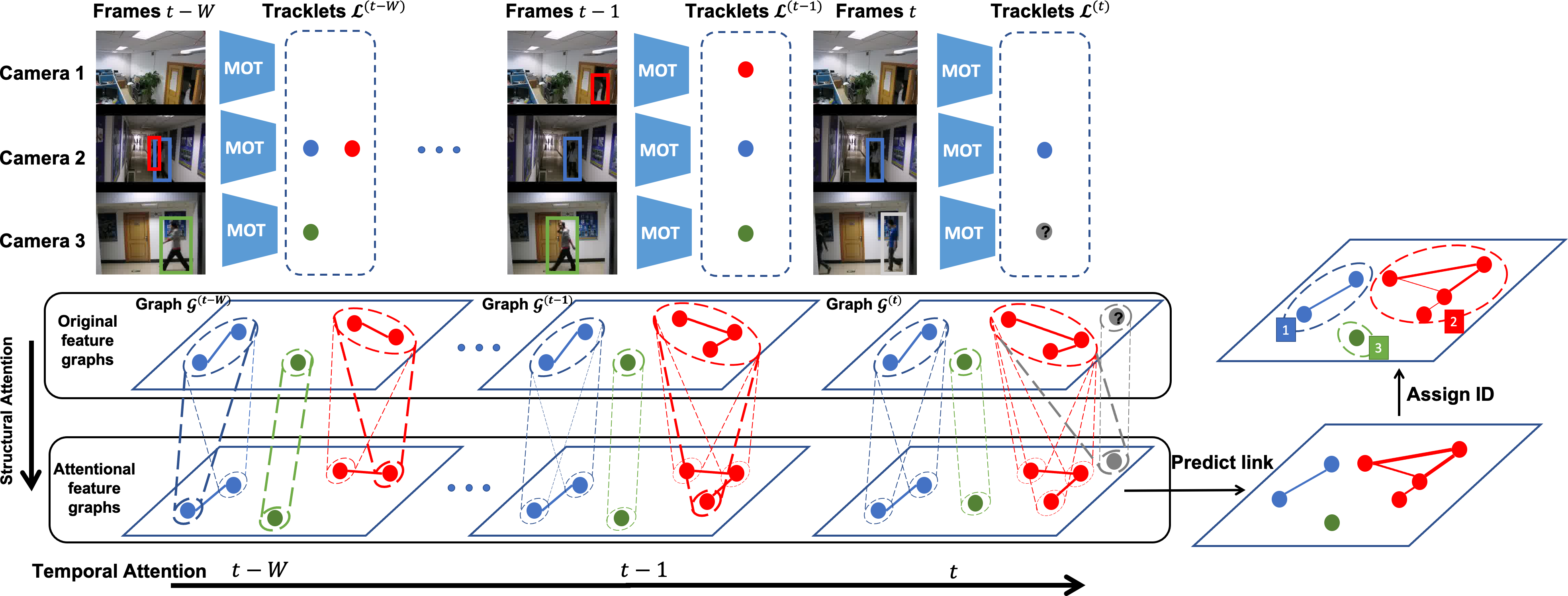}
    \caption{The proposed DyGLIP framework. Connected components, grouped by dash boundaries, represent known tracklets of unique objects in a multi-camera system. The gray node with a question mark is a new subject that has appeared in camera 3 at time step $t$. Such a  node is successively added to our graph over time, its transformed features are computed by attending to existing nodes in the graph, and then its connections are predicted using structural and temporal attention feature embedding from graphs up to $t - W$ steps.}
    \label{fig:framework}
    \vspace{-4mm}
\end{figure*}

\section{The Proposed \textit{DyGLIP} Method}

In this section, we first briefly define the MC-MOT problem in Subsection \ref{sec:problem_formulation}. It is then re-formulated as a construction of a dynamic graph with an ID assignment problem in Subsection \ref{sec:dyn_graph}.
A new attention mechanism is presented to enhance the robustness of the associated features for each node in the proposed graph in 
Subsection \ref{sec:attn_dyn_graph}.
Then ID assignment problem is solved via link prediction as in Subsection \ref{sec:link_prediction} to connect new nodes with existing nodes. Finally, we present the model learning in Section \ref{sec:model_learning} and some analysis on model complexity in Subsection \ref{sec:model_complexity}.

\subsection{Problem Formulation}
\label{sec:problem_formulation}
In MC-MOT, it is assumed that the environment is continuously monitored by $C$ cameras, denoted by the set $\cC=\{c_1,\dots, c_C\}$. In contrast to some prior methods \cite{you2020realtime, Chen_2020_CVPR} that require certain overlapping ratios between the cameras, DyGLIP can gracefully handle both overlapping and non-overlapping scenarios.
Similar to prior MC-MOT methods \cite{he2020multi, Qian_2020_CVPR_Workshops}, the task of tracking in each camera is assumed to be performed by an off-the-shelf single-camera MOT tracker. We choose DeepSORT \cite{Wojke2017simple} in this work, but it can be simply replaced by any other MOT trackers.
%
\begin{comment}
\textcolor{red}{We use $\bo_i$ to denote a tracked object where the subscript $i$ represents its assigned ID, and the set $\cO^{(t)}$ gathers all the tracked objects at time $t$. Note that the cardinality $|\cO^{(t)}|$ varies over time due to the fact that new objects could appear, or existing objects could move out of the tracking region, hence they are removed from $\cO$. 
For a particular object $\bo_i$, we denote by $\cL^{(t)}_i$ its associated trajectory (tracklet) at time step $t$. If an object $\bo_i$ appears at time step $t_0$, $\cL^{(t)}_i$ can be written as $\cL_i^{(t)} = \{\bl^{(t_0)}_i, \dots, \bl^{(t)}_i \}$, where $\bl^{(t)}_i$ stores the features of the object $\bo_i$ at time $t$. Depending on the application, the vector $\bl^{(t)}_i$ can be chosen to contain richer information about the object. In this work, we choose $\bl^{(t)}_i$ to be the feature vector obtained from the underlying MOT tracker.}
\end{comment}
%
At time step $t$, we obtain a set of local tracklets $\cL_c^{(t)} = \{\bl^{(t)}_j\}$ provided by each single-camera MOT tracker, where each $\bl^{(t)}_j$ is a feature vector. Sometimes it is referred to as re-id features in the literature. Note that the set $\cL_c^{(t)}$ may contain tracklets of new objects, and the global IDs of the tracklets are unknown. Moreover, one object could associate with multiple local tracklets across camera views. The aim of data association is, therefore, to assign the proper global IDs for the local tracklets in $\cL_c^{(t)}$.

In practice, given an unassigned tracklet $\bl^{(t)}_j$, most MC-MOT algorithms rely on the information in $\bl^{(t)}_j$ to find its correlation to the set of known tracklets $\cup_{c=1}^{C} \cup_{k=1}^{t-1} \cL_c^{(k)}$ obtained from previous time steps. In this work, we show that solely using information from the feature vectors $\bl^{(t)}_j$ could lead to a wrong data association. Our proposed method, on the other hand, uses the set of obtained tracklets and embeds them into a dynamic graph model. The graph is then equipped with structural and temporal attention, offering us a richer tracklet representation. 

\subsection{Dynamic Graph Formulation}
\label{sec:dyn_graph}

We first introduce our graph formulation (See Fig. \ref{fig:framework}). At a particular time step $t$, we construct a graph $\cG^{(t)} = (\cV^{(t)}, \cE^{(t)})$, where the vertex set $\cV_t$ contains \emph{all the tracklets} tracked up to time $t$. 
Unlike existing works that build a fixed graph and conduct relevant data association algorithms, we maintain a \emph{dynamic graph} during our tracking process, which is a key novelty of our work. More specifically, at each time step $t$, new vertices are added into our graph, hence our vertices are growing over time, i.e., $\cV^{(t)} = \cV^{(t-1)} \cup \cN^{(t)}$, where we use $\cN^{(t)}$ to denote the set of new vertices, i.e., new tracklets, obtained from the MOT tracker in each camera. 
Before the data association step, the connection between these new vertices to the vertices of $\cV^{(t-1)}$ is not determined. 

We denote $f(v)$ as the feature vector associate with a node $v \in \cV^{(t)}$. Given two nodes $v_i$ and $v_j$, an edge exists that links the two vertices if these two tracklets represent the same object. From our experiments, we choose $f(v)$ to be the re-id feature of the tracklet associated with node $v$. 
In the following, we will discuss how the attention mechanism can be utilized to extract the embedding features for each node dynamically and how the robust connections can be established using link prediction.

\subsection{Dynamic Graph with Self-Attention Module}
\label{sec:attn_dyn_graph}
We now introduce the self-attention module in our proposed dynamic graph and its building blocks, i.e., graph structural and temporal attention layers. 
Although using attention for dynamic graph has been considered in the literature \cite{xu2020inductive,velivckovic2017graph,sankar2020dysat}, their methods cannot be straightforwardly applied to our problem.
Inspired by these works, we propose a novel self-attention mechanism (see Fig. \ref{fig:framework}) for our model. 
The goal of these attention layers is to attend inter-cameras and intra-cameras information, which allows our model to capture variations between tracklets, as well as attend the temporal information over multiple time steps.
\subsubsection{Graph Structural Attention Layer}
Our attention layer takes into account not only the provided embedding features but also the camera information. In other words, the structural attention layer (SAL) takes the concatenation of node embeddings or features, i.e., $f(v) \in \mathbb{R}^{D_F}$, and its camera positional encoding, i.e., $ \mathbf{c}_v \in \mathbb{R}^{D_C}$, as the input, $ \mathbf{e}_v = \left\{ f(v) || \mathbf{c}_v \right\} \in \mathbb{R}^{D_E}$, where $D_E = D_F + D_C$.

Given a set of camera-aware node features from a graph $\cG ^ {(t)}$, $ \mathbf{e}^{(t)} = \{ \mathbf{e}^t_1, \cdots, \mathbf{e}^t_N \}$, where $N = |\cV^{(t)}|$, our structural attention layer provides a new set of node features $\mathbf{h}^{(t)} = \{ \mathbf{h}^t_1, \cdots, \mathbf{h}^t_N \}, \mathbf{h}_v \in \mathbb{R}^{D_{H}}$, as the output.
The learning of self-attention features can be stabilized with multi-head attention, where the input features are transformed by $L$ independent transformation and their transformed features are concatenated as in Eqn. \eqref{eqn:str_att_layer}.
\begin{equation} \label{eqn:str_att_layer}
\small
    \mathbf{h}^{t}_{v_i} = \overset{L}{\underset{l=1}{\text{Concat}}} \left[ \sigma \left( \sum_{v_j \in \cV^{(t)}} \alpha^l_{ij} \text{conv}^l_{1 \times 1} \left( \mathbf{e}_{v_j}^{t} \right)  \right) \right]
\end{equation}
where $\text{conv}^l_{1 \times 1}$ is a 1D convolutional layer with kernel size $1 \times 1$, $\sigma (\cdot)$ is a non-linear activation function, i.e., LeakyRELU, and $\alpha^l_{ij}$ are the attention coefficients for the $l$-th attention head as in Eqn. \eqref{eqn:coefficents}.
\begin{equation} \label{eqn:coefficents}
\footnotesize
    \alpha^l_{ij} = \frac{ \exp \left( \sigma \left( \mathbf{W}_{ij}^T \left[ \text{conv}^l_{1 \times 1} ( \mathbf{e}_{v_i}^{t} ) \Vert \text{conv}^l_{1 \times 1} ( \mathbf{e}_{v_j}^{t} ) \right] \right) \right) } { \sum_{v_k \in \cV^{(t)} } \exp \left( \sigma \left( \mathbf{W}_{kj}^T \left[ \text{conv}^l_{1 \times 1} ( \mathbf{e}_{v_k}^{t} ) \Vert \text{conv}^l_{1 \times 1} ( \mathbf{e}_{v_j}^{t} ) \right] \right) \right) }
\end{equation}
where $\Vert$ is the feature vector concatenation operation, $\mathbf{W}_{ij}^T \in \mathbb{R}^{D_E \times D_{H}} $ is the shared weight of the transformation applied to edge $ (v_i, v_j) $ in the graph at time step $t$. 
Those normalized (by softmax) attention coefficients $\alpha^l_{ij}$ indicate the impact of node $v_j$'s features to node $v_i$. To incorporate graph structure, we employ sparse adjacent matrices to compute $\alpha^l_{ij}$ for nodes $j$ within its neighborhood of node $i$ while ignoring others. 
This layer can be stacked to create multiple structural attention layers and applied independently for each graph $\cG^{(t)}$ to capture the local structure of a node at each time step.
In the next section, we will introduce how to attend to the graph dynamic's evolution or development across multiple time steps by using the temporal attention layer, which takes the node representation output from the structural attention layers.

\subsubsection{Graph Temporal Attention Layer}

A temporal attention layer (TAL) is designed to capture the temporal evolution scheme in terms of links between nodes in a set of dynamic graphs. 
This layer takes a set of output representations or features from structural attention layers at different time steps and timestamps as inputs and provides the time-aware features for each node at each time step $t$. 
In particular, for each node $v \in \cV^{(t)}$, we combine timestamp position encoding ($ \mathbf{p}_v \in \mathbb{R}^{D_H} $) with the output from the SAL to obtain an order-aware sequence of input features for TAL as, $ \mathbf{x}_v = \left[ \mathbf{h}^1_v + \mathbf{p}_v^1, \mathbf{h}^2_v + \mathbf{p}_v^2, \cdots, \mathbf{h}^W_v + \mathbf{p}_v^W \right]$, where $W$ is temporal window size, i.e., $W$ = 3 in our experiments, for each node $v \in \cV$.
We stacked multiple time step together as matrices $ \mathbf{X} \in \mathbb{R}^{W \times D_H}$. 
The $l$-th output of the multi-head TAL are defined using scaled dot-product attention as in Eq. \eqref{eqn:tem_att_layer}
\begin{equation} \label{eqn:tem_att_layer}
\small
    \mathbf{z}^{(l)}_e = \text{attn}^{(l)} ( \mathbf{Q}, \mathbf{K}, \mathbf{V} ) = \text{softmax} \left( \frac{\mathbf{Q} \mathbf{K}^T}{\sqrt{D_Z}} + \mathbf{M} \right) \mathbf{V}
\end{equation}
where $l = 1, \cdots, L$, with $L$ is the number of heads, $\mathbf{Q} = \mathbf{X} \mathbf{W}_Q $, $\mathbf{K} = \mathbf{X} \mathbf{W}_K $, and $\mathbf{V} = \mathbf{X} \mathbf{W}_V$ are the ``queries", ``key" and ``values" transformed features by linear projection matrices $\mathbf{W}_Q \in \mathbb{R}^{D_H \times D_Z}$, $\mathbf{W}_K \in \mathbb{R}^{D_H \times D_Z}$ and $\mathbf{W}_V \in \mathbb{R}^{D_H \times D_Z}$, respectively, as defined in \cite{vaswani2017attention}, $D_Z$ is the output feature dimension and $ \mathbf{M} \in \mathbb{R}^{W \times W}$ is the mask matrix where each element $ \mathbf{M} \in \{ -\infty, 0 \} $.
We employ such masked attention to allow backward attention where each time step attends over all previous time steps. Thus, to set a zero attention weight, we assign $\mathbf{M}_{ij} = -\infty$, where $i > j$, otherwise $\mathbf{M}_{ij} = 0$.
Similarly, we can create multiple stacked temporal attention layers by stacking temporal attention layers together.
Then the multi-head output of the final layer will be concatenated and passed to a feed-foward neural network to capture non-linear interactions between the transformed features to provide the final set of node embeddings $\{ \mathbf{e'}^{(1)}, \mathbf{e'}^{(2)}, \cdots, \mathbf{e'}^{(W)} \}$ as in \cite{xu2020inductive}. 

\subsection{Link Prediction}
\label{sec:link_prediction}

This section describes how we compute the probability of having a connection/link between two nodes and how we learn a link classifier jointly with the attention module to obtain the transformed features. In this way, we can guarantee that the self-attention module and the classifier provide the best link prediction accuracy.
Given transformed features of a pair of nodes ($  \mathbf{e'}_{v_i}^{(t)} $ and $ \mathbf{e'}_{v_j}^{(t)} $), we compute the features or measurement that represent the similarity between those two nodes, and then it will be used as input for the classifier. The classifier provides a probability score $s \in [0, 1]$. The higher the score is, the more likely the two nodes are linked.
We try with two different measurements and classifiers in Section \ref{ssec:ablation_study}, i.e., cosine distance by computing dot product of two feature vectors with Sigmoid as classifier and the Hadamard operator ( $ \mathbf{e'}_{v_i}^{(t)} \odot \mathbf{e'}_{v_j}^{(t)} $ ) with a fully connected layer and a softmax layer as classifier.

\subsection{Model Learning}
\label{sec:model_learning}

To train the proposed DyGLIP framework, our objective function is to learn representations capturing both structural and temporal information from dynamic graphs as well as to predict possible links between two arbitrary nodes in the graph using the learned representations.
We first use a binary cross-entropy loss function to enforce nodes within a connected component to have similar feature embeddings. Then a classifier loss function to ensure classify two nodes based on measurement features as having a link or not.
\vspace{-2mm}
\begin{equation}
\footnotesize
\begin{split}
    \mathcal{L}(v_i) = &\sum_t^T \left( \sum_{v_j \in \cN_b^{(t)}(v_i)}  -\log 
    \left( \sigma \left( < e'_{v_i} , e'_{v_j} > \right) \right) \right.  \\ 
    &\left. - w_g \sum_{v_k \in \cN_g^{(t)}(v_i)} \log \left( 1 - \sigma \left( < e'_{v_i} , e'_{v_k} > \right) \right)  \right. \\ 
    &\left. + \sum_{v_j \in \cN_a^{(t)}(v_i)} \mathcal{L}_c (v_i, v_j) \right) 
\end{split}
\end{equation}
where $ <\cdot> $ is the inner production between two vectors, $\sigma$ is Sigmoid activation function, $\cN_b^{(t)}(v_i)$ is the set of fixed-length random walk neighbor nodes of $v_i$ at time step $t$, $\cN_g^{(t)}(v_i)$ is a negative samples of $v_i$ for time step $t$,  $\cN_a^{(t)}(v_i) = \cN_b^{(t)}(v_i) \cup \cN_g^{(t)}(v_i)$ and $w_g$, negative sampling ratio, is an adjustable hyper-parameter to balance the positive and negative samples. $\mathcal{L}_c$ is the loss for classifier.

\subsection{Algorithmic Complexity}
\label{sec:model_complexity}
This section briefly discusses the computational complexity of our approach. 
We assume the network structure is fixed; hence the dimensions of the embedding feature vectors and layers are constant numbers. Therefore, the complexity of one network pass is constant, i.e., $O(1)$. Let us consider the graph $\cG^{(t)} = (\cV^{(t)}, \cE^{(t)})$ at the $t$-th time step. It can be observed that the overall complexity of our model depends on the structural and temporal attention modules.

\vspace{2mm}
\noindent
\textbf{Graph Structural Attention Module.} 
From Eqn.~\eqref{eqn:str_att_layer}, the time complexity to compute $\bh^{t}_{v_i}$ depends on the number of attention coefficients $\alpha^t_{ij}$. Due to the fixed structure assumption, the computation of $\alpha^t_{ij}$ in Eqn.~\eqref{eqn:coefficents} is $O(1)$. Thus, the overall time complexity of the Eqn. \eqref{eqn:coefficents} is $O(|\cV| + |\cE|)$. 

\begin{figure}[!t] 
    \centering
    \includegraphics[width=0.6\columnwidth]{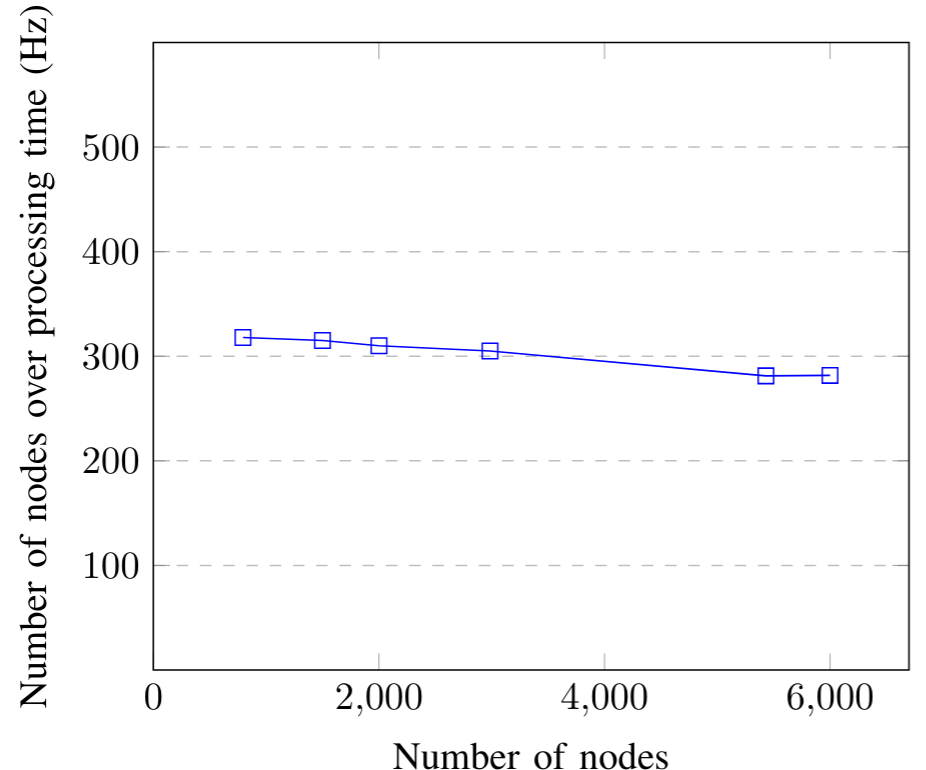}
    \caption{The performance of DyGLIP in terms of Hz w.r.t. number of nodes/tracks.}\label{fig:Hz}
\end{figure}

\vspace{2mm}
\noindent
\textbf{Graph Temporal Attention Module} The time complexity of this module is equivalent to the time complexity of Eqn. \eqref{eqn:tem_att_layer}. We ignore the mask $\mathbf{M}$ in Eqn. \eqref{eqn:tem_att_layer}, the time complexity of Eqn. \eqref{eqn:tem_att_layer} is dependent on the matrix multiplication between $\mathbf{Q}, \mathbf{K}$ and  $\mathbf{V}$, i.e., $\operatorname{softmax}(\frac{\mathbf{Q}\mathbf{K}^T}{\sqrt{D_Z}})\mathbf{V}$. Mathematically, the time complexity of operation $\operatorname{softmax}(\frac{\mathbf{Q}\mathbf{K}^T}{\sqrt{D_Z}})\mathbf{V}$ is $O(W D_Z ^2 + W^2 D_Z)$
($W$ is the window time size). However, $D_Z$ is a constant number; hence, the time complexity will be equivalent to $O(W^2)$.
Consequently, the time complexity of our network is grown with respect to the size of graph $\cG^{(t)}$ and the squared window time size $W^2$, i.e., $O(|\cV| + |\cE| + W^2)$. As $W$ can be chosen to be fixed during the operation, our network's overall complexity grows with the number of tracks. Our theory 
aligns with the empirical results shown in Fig. \ref{fig:Hz}, where the number of nodes can be processed per second (Hz) slightly drops when the number of nodes in the graph increases from 2K to 6K.

\section{Experimental Results}

In this section, we conduct the experiments to demonstrate the benefits of attention modules, link regression and feature extraction for each nodes in terms of link prediction accuracy, i.e., Accuracy Under Curve (AUC) \cite{grover2016node2vec} and ID assignment accuracy, i.e., \textbf{1. CLEAR MOT metrics} \cite{bernardin2008evaluating} including MOTA, MOTP, ID Switch (IDS); \textbf{2. ID scores} \cite{ristani2016performance} including F1 (IDF1), Precision (IDP), Recall (IDR); and \textbf{3. Multi-camera Tracking Accuracy} (MCTA) \cite{chen2016equalized}. 
We also compare with other state-of-the-art approaches for both overlapping and non-overlapping FOVs on human and car multi-camera tracking datasets.

\begin{table}[!t]
\footnotesize
    \centering
    \begin{tabular}{|c|c|c|c|c|}
    \hline
         \textbf{Datasets} & \textbf{\#frames} & \textbf{\#box} & \textbf{\#ID} & \textbf{\#Camera}   \\
         \hline
         \textbf{HDA} & 75,207 & 64,028 & 85 & 13 \\
         \textbf{SAIVT-Softbio} & 41,979 & 64,472 & 152 & 8 \\
         \textbf{PRW} & 11,816 & 34,304 & 932 & 6 \\
         \textbf{MARS} & 20,715 & 509,914 & 625 & 6 \\
         \hline
         \textbf{Total} & \textbf{150K} & \textbf{672K} & \textbf{1.8K} & \textbf{33} \\
         \hline
    \end{tabular}
    \caption{Statistics of the joint training set.}
    \label{tab:summary_datasets}
    \vspace{-4mm}
\end{table}

\subsection{Datasets}

Experiments conduct on small-scale datasets may provide biased results and may not valid when applying to large-scale datasets. In addition, Duke-MTMC \cite{ristani2016performance}, a large-scale datasets, was commonly used in MC-MOT research community is no longer publicly available. Thus, we introduce a large-scale dataset\footnote{This dataset will be publicly available.} by putting together eight publicly available datasets on Multi-camera settings, four for training and four for testing. Those datasets including \textbf{HDA Person} \cite{nambiar2014multi}, \textbf{SAIVT-Softbio} \cite{quteprints53437}, 
\textbf{PRW} \cite{zheng2016person}, \textbf{MARS} \cite{zheng2016mars}, \textbf{PETS09} \cite{PETS09}, \textbf{CAMPUS} \cite{xu2016multi}, \textbf{EPFL} \cite{EPFL} and \textbf{MCT} \cite{chen2016equalized}. 
Training subsets of HDA Person, SAIVT-Softbio, PRW, and MARS datasets are gathered to form the joint training set.
Since our proposed DyGLIP does not require full video frames for training, we can use large-scale ReID datasets, e.g., MARS, and build dynamic graph and node's features based on the provided information for each bounding box, i.e., the ID of the pedestrian, camera, tracklet ID and frame number.
We can also train on datasets without continuous frames, e.g., PRW, it still shows the evolution of dynamic graph as tracked subjects change over time.
After combining the above datasets, we obtain a training set contains a total of 150K frames, with a total of 1.8K unique ID (see Table \ref{tab:summary_datasets} for more details).
A small subset of this joint training set is used as a validation set for our ablation studies in Subsection \ref{ssec:ablation_study}.
We use four public benchmark datasets, including PETS09, CAMPUS, EPFL, and MCT, for performance evaluations. We also train and test on car datasets, i.e., CityFlow \cite{tang2019cityflow} from AI City Challenge 2020 \cite{Naphade20AIC20}.

\subsection{Experimental Setup}

We conduct MC-MOT experiments by learning dynamic graph representation with two time steps $ \{\cG^{(t - 2)}, \cG^{(t - 1)} \} $ to predict/assign ID for new nodes in $ \cG^{(t)} $ at time step $t$. Thus, training data are split into mini-batch of a chunk size of 3 and employed a mini-batch gradient descent with the Adam optimizer to learn all the parameters of the attention module and the classifier.
The attention module is implemented in Tensorflow \cite{abadi2016tensorflow}. We use two SALs with four heads, each computing 128 features (layer sizes of 512) and two TALs with 16 heads, each computing eight features (output sizes of 128), given the input raw (unattended) feature dimension of 512. The model is trained with a maximum of 100 epochs with a batch size of 512 chunks. Note that we apply some padding to combine those chunks (with a different number of nodes) into a batch for training. We choose the best performing model on the validation set for evaluation on four benchmark datasets. 

\subsection{Ablation Studies}
\label{ssec:ablation_study}
This section presents several deeper studies to analyze our proposed model and justify our competitive performance presented in the following sections. More specifically, this section aims to demonstrate the following appealing properties of the proposed method: (1) \textit{\textbf{Better feature representations}}, even in severe changes in lighting conditions between the cameras; and (2) \textit{\textbf{Recovery of correct representations for objects}} that lost tracks during camera transitions.
In addition, we also conduct several other studies to evaluate the role of link regression and the influence of initial node embedding feature choices.

\begin{figure*}[!t]
    \centering
    \includegraphics[width=1.7\columnwidth]{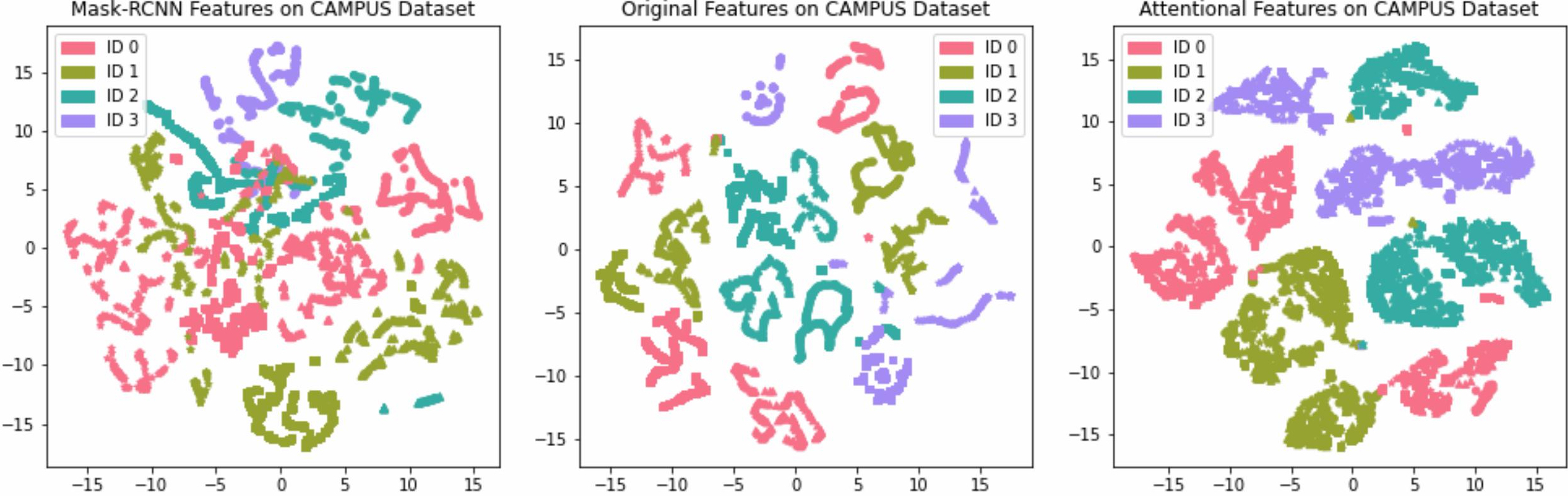}%
    \caption{Node embedding in t-SNE space (a) Features from detector \cite{he2017mask}. (b) Original features from ReID model \cite{zhou2019osnet}. (c) Transformed features with attention. Same color indicating same subject and same symbol indicating same camera. Each node corresponding to a time step in the video sequences. (\textbf{Best viewed in color})}
    \label{fig:feature_attention}
    \vspace{-4mm}
\end{figure*}

\begin{figure}[!t]
    \centering
    \includegraphics[width=0.7\columnwidth]{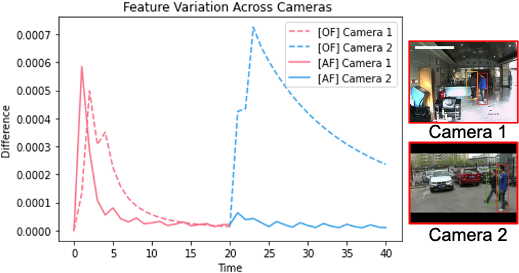}
    \caption{Feature variation across cameras. OF and AF denotes for original features and attentional features, respectively.}
    \label{fig:feature_varriation}
    \vspace{-4mm}
\end{figure}

% \begin{figure}
%     \centering
%     \subfloat{%
%     \includegraphics[width=1.8\columnwidth]{figs/final_campus_garden1_features.jpg}%

%     \label{fig:feature_attention}
%         }%
%     \hfill%
%     \subfloat{%
%         \includegraphics[width=0.93\linewidth]{figs/Failed_cases.jpg}%
%         \label{fig:failed_detection_case}%
%         }%
%     \caption{Node embedding in t-SNE space (a) Features from detector \cite{he2017mask}. (b) Original features from ReID model \cite{zhou2019osnet}. (c) Transformed features with attention. Same color indicating same subject and same symbol indicating same camera. Each node corresponding to a time step in the video sequences. (\textbf{Best viewed in color})}
%     % \includegraphics[width=1\linewidth]{figs/overview.jpg}
%     % \caption{Samples of multi-view captured by multi-cameras setup on a vehicle from nuScenes dataset \cite{caesar2020nuscenes}}
%     % \label{fig:sample_nuscenes}
% \end{figure}

\begin{table}[!t]
\footnotesize
    \centering
    \begin{tabular}{|c|c|c|c|c|}
    \hline
        \textbf{Features} &  \textbf{ID F1} (\%) $\uparrow$ & \textbf{IDP} (\%) $\uparrow$ & \textbf{IDR} (\%) $\uparrow$ & \textbf{IDS} $\downarrow$  \\
         \hline
         DyGLIP & \textbf{56.2} & \textbf{59.5} & \textbf{56.2} & \textbf{44} \\
         $-$ att & 39 & 50.1 & 36.5 & 135 \\
         
         \hline
    \end{tabular}
    \caption{MOT metrics comparison between DyGLIP (with attention modules) and $-$ att (without attention modules).}
    \label{tab:mot_ablation}
\end{table}

\begin{table}[!t]
\footnotesize
    \centering
    \begin{tabular}{|c|c|c|}
    \hline
         \textbf{Method} & \textbf{Val AUC} & \textbf{Test AUC} \\
         \hline
         Link Regression with Attention & \textbf{97.79} \% & \textbf{97.48} \% \\ 
         Link Regression w/o Attention & 84.23 \% & 87.36 \% \\
         \hline
    \end{tabular}
    \caption{Accuracy Under the Curve (AUC) comparison between using and without using attention modules.}
    \label{tab:assign_id}
\end{table}

\textbf{Better Representations}
It is well-known in most tracking applications that the data association task's accuracy is dependent mainly on the quality of the underlying feature representations. In other words, one expects that features representing the same object (from different cameras) during its trajectory must form a cluster in the feature space. While other methods use the features produced by MOT trackers, we will show in this section that our dynamic graph with the structural and temporal attention mechanism offers better representations for each node in the graph. 
We use a subset of frames from the same video for training and validating our method and use the ones in another video for testing. Note that our testing frames cover multiple objects, where each object has multiple transitions over different cameras. The features produced by our method for all the nodes are plotted (using t-SNE embedding~\cite{hinton2002stochastic}) in Fig.~\ref{fig:feature_attention} (c), while the original node features obtained from MOT trackers are also plotted in Fig.~\ref{fig:feature_attention} (b). In this figure, features produced by DyGLIP form better clusters than the original features. Furthermore, to quantitatively justify the role of attention, Tables~\ref{tab:mot_ablation} and \ref{tab:assign_id} show the results with and without attention, where the attention mechanism significantly improves the performance.
Fig. \ref{fig:feature_varriation} illustrates how feature embedding of a subject/node change-over-time. Especially when the subject moves from one camera to another camera, the original features (OF) change with a large margin while the transformed features (AF) are quite stable. We measure the average between the cluster centroid and all the corresponding node features up to the time step $t$.
\begin{table}
\footnotesize
    \centering
    \begin{tabular}{|c|c|c|c|}
    \hline
         \textbf{Metrics} & \textbf{Classifier} & \textbf{Val AUC} & \textbf{Test AUC} \\
         \hline
         Cosine distance & Sigmoid & 81.9 \% & 68.1 \% \\
         Hadamard product & SM & \textbf{97.79} \% & \textbf{97.48} \%  \\
         \hline
    \end{tabular}
    \caption{Area Under Curve (AUC) between various feature metrics and link classifiers, i.e., Sigmoid and Softmax (SM).}
    \label{tab:link_classify}
\end{table}

\begin{figure*}[!t]
    \vspace{-3mm}
    \centering
    \includegraphics[width=0.7\textwidth]{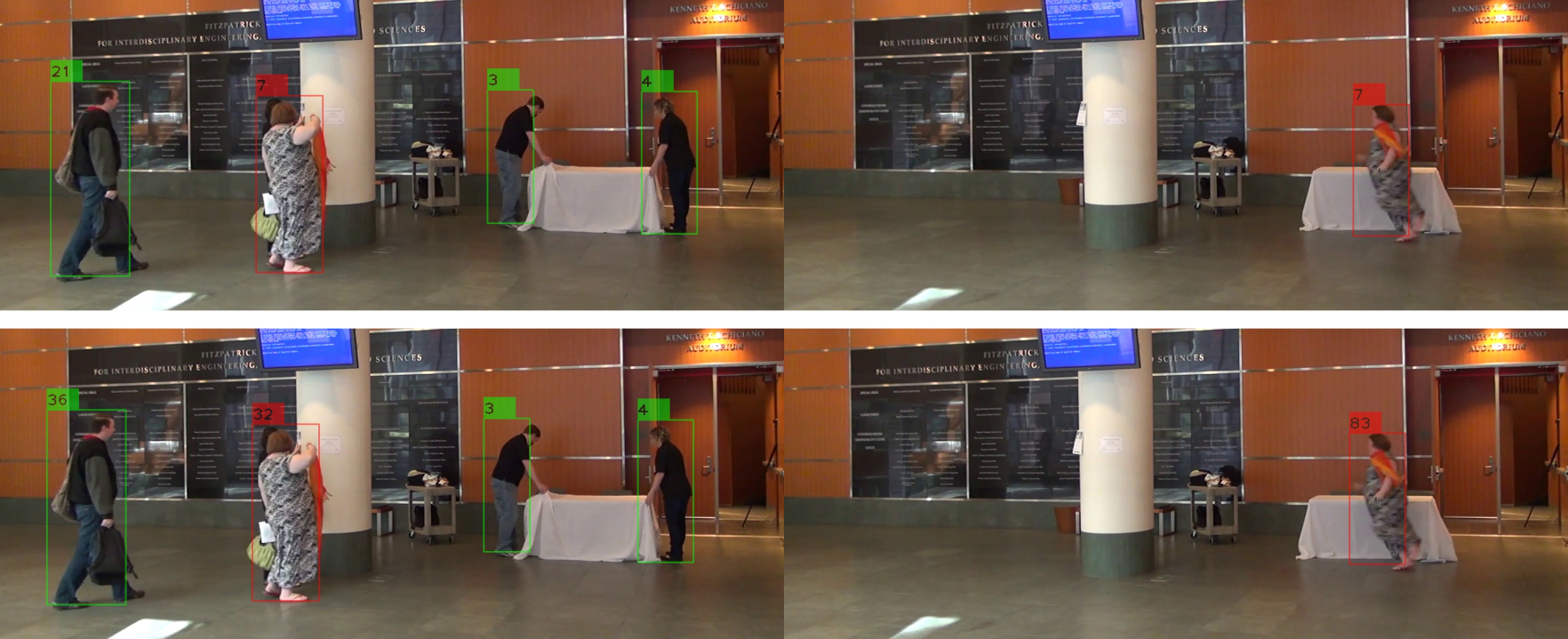}
    \caption{Our proposed method (up) corrects a negative matched case caused by a short-memorized MOT system (down). Determined ID is highlighted by the red bounding box.}
    \label{fig:recover_cluster}
\end{figure*}

\begin{table}
\footnotesize
    \centering
    \begin{tabular}{|c|c|c|}
    \hline
         \textbf{Features} & \textbf{Val AUC} & \textbf{Test AUC} \\
         \hline
         ReID \cite{zhou2019osnet} & \textbf{97.79} \% & \textbf{97.48} \% \\
         Detector \cite{he2017mask} & 73.95 \% & 66.75 \% \\
         \hline
    \end{tabular}
    \caption{Area Under Curve (AUC) comparison between various initial features for each node in the proposed graph, i.e., ReID features or features from human detector.}
    \label{tab:feature_reid}
    \vspace{-4mm}
\end{table}
\textbf{Recover from Fragmented Tracks in single-camera.}
We demonstrate the ability to recover from assignment error in the previous steps using cluster of nodes, i.e., a set of nodes that form a connected component in the graph. Fig. \ref{fig:recover_cluster} illustrates our proposed DyGLIP can recover single-camera MOT mistake on-the-fly.

\textbf{Roles of Link Regression.}
We study the effects of different metrics and classifiers to predict the probability of having an edge connecting two nodes given their embedding features indicating they are in the same ID in Table \ref{tab:link_classify}. Our objective function optimizes the inner product that helps the model achieve much better AUC with Hadamard product metric and softmax classifier.
 
\begin{table}[!t]
\footnotesize
    \centering
    \begin{tabular}{|c|c|c|c|}
    \hline
    \textbf{Sequence} & \textbf{Method} & \textbf{MOTA} (\%) $\uparrow$ & \textbf{MOTP} (\%) $\uparrow$ \\ 
    \hline
    S2L1         & KSP \cite{berclaz2011multiple} & 80 & 57 \\ 
                 & B\&P \cite{rosenhahn2012branch} & 72 & 53 \\ 
                 & HCT \cite{xu2016multi} & 89 & 73 \\ 
                 & TRACTA \cite{he2020multi} & 87.5 & 79.2 \\ 
                 \cline{2-4}
                 & \textbf{DyGLIP} & \textbf{93.5} & \textbf{94.7} \\ 
    \hline
    \end{tabular}
    \caption{Evaluation results on PETS09 dataset.}
    \label{tab:PETS09_results}
    \vspace{-4mm}
\end{table}
\begin{table}[!t]
\footnotesize
    \centering
    \resizebox{\linewidth}{!}{%
    \begin{tabular}{|c|c|c|c|}
    \hline
    \textbf{Sequence} & \textbf{Method} & \textbf{MOTA} (\%) $\uparrow$ & \textbf{MOTP} (\%) $\uparrow$ \\ 
    \hline
    Passageway   & KSP \cite{berclaz2011multiple} & 40 & 57 \\ 
                 & HCT \cite{xu2016multi} & 44 & 71 \\ 
                 & TRACTA \cite{he2020multi} & 52.1 & 77.5 \\ 
                 \cline{2-4}
                 & \textbf{DyGLIP} & \textbf{70.4} & \textbf{97.2} \\
    \hline    
    Basketball   & KSP \cite{berclaz2011multiple} & 56 & 54 \\ 
                 & HCT \cite{xu2016multi} & 60 & 68 \\ 
                 & TRACTA \cite{he2020multi} & 64.3 & 72.5 \\ 
                 \cline{2-4}
                 & \textbf{DyGLIP} & \textbf{66.3} & \textbf{89.5} \\ 
                 \hline
    \end{tabular}}
    \caption{Evaluation results on EPFL dataset.}
    \label{tab:epfl_results}
    \vspace{-4mm}
\end{table}
\begin{table}[!t]
    \footnotesize
    \centering
    \resizebox{\linewidth}{!}{%
    \begin{tabular}{|c|c|c|c|c|c|}
    \hline
    \textbf{Seq} & \textbf{Method} & \textbf{MOTA} $\uparrow$ & \textbf{MOTP} $\uparrow$ & \textbf{MT} $\uparrow$ & \textbf{ML} $\downarrow$  \\        \hline
    \parbox[t]{2mm}{\multirow{4}{*}{\rotatebox[origin=c]{90}{Garden 1}}}     
                 & HCT \cite{xu2016multi} & 49\% & 71.9\% & 31.3\% & 6.3\% \\
                 & STP \cite{xu2017cross} & 57\% & 75\% & -- & -- \\
                 & TRACTA \cite{he2020multi} & 58.5\% & 74.3\% & 30.6\% & 1.6\% \\
                 \cline{2-6}
                 & \textbf{DyGLIP} & \textbf{71.2}\% & \textbf{91.6}\% & \textbf{31.3}\% & \textbf{0.0}\% \\
    \hline
    \parbox[t]{2mm}{\multirow{4}{*}{\rotatebox[origin=c]{90}{Garden 2}}}
                 & HCT \cite{xu2016multi} & 25.8\% & 71.6\% & 33.3\% & 11.1\% \\
                 & STP \cite{xu2017cross} & 30\% & 75\% & -- & -- \\
                 & TRACTA \cite{he2020multi} & 35.5\% & 75.3\% & 16.9\% & 11.3\% \\
                 \cline{2-6}
                 & \textbf{DyGLIP} & \textbf{87.0}\% & \textbf{98.4}\% & \textbf{66.67}\% & \textbf{0.0}\% \\
    \hline
    \parbox[t]{2mm}{\multirow{4}{*}{\rotatebox[origin=c]{90}{Auditorium}}}
                 & HCT \cite{xu2016multi} & 20.6\% & 69.2\% & 33.3\% & 11.1\% \\
                 & STP \cite{xu2017cross} & 24\% & 72\% & -- & -- \\
                 & TRACTA \cite{he2020multi} & 33.7\% & 73.1\% & 37.3\% & 20.9\% \\
                 \cline{2-6}
                 & \textbf{DyGLIP} & \textbf{96.7}\% & \textbf{99.5}\% & \textbf{95.24}\% & \textbf{0.0}\% \\
    \hline
    \parbox[t]{2mm}{\multirow{4}{*}{\rotatebox[origin=c]{90}{Parkinglot}}}   
                 & HCT \cite{xu2016multi} & 24.1\% & 66.2\% & 6.7\% & 26.6\% \\
                 & STP \cite{xu2017cross} & 28\% & 68\% & -- & -- \\
                 & TRACTA \cite{he2020multi} & 39.4\% & 74.9\% & 15.5\% & 10.3\% \\
                 \cline{2-6}
                 & \textbf{DyGLIP} & \textbf{72.8}\% & \textbf{98.6}\% & \textbf{26.67}\% & \textbf{0.0}\% \\
                 \hline
    \end{tabular}}
    \caption{Evaluation results on CAMPUS dataset.}
    \vspace{-4mm}
    \label{tab:campus_results}
\end{table}
\begin{table*}[!t]
\centering
\footnotesize
    \begin{tabular}{|c|c|c|c|c|c|c|c|}
    \hline
         \textbf{Subset} & \textbf{Method} & \textbf{MCTA} (\%) $\uparrow$ & \textbf{MOTA} (\%) $\uparrow$ & \textbf{MOTP} (\%) $\uparrow$ & \textbf{Precision} (\%) $\uparrow$ & \textbf{Recall} (\%) $\uparrow$ & \textbf{IDS} $\downarrow$  \\
         \hline
         Dataset1 & EGM \cite{chen2016equalized} & 41.2 & 59.4 & 68.0 & 79.7 & 59.2 & 1888 \\
            & RAC \cite{cai2014exploring} & 59.5 & 92.6 & 64.6 & 69.2 & 60.6 & 154 \\
            & ICLM \cite{lee2017online} & 61.2 & 87.3 & 68.1 & 77.2 & 60.9 & 112 \\
            & TRACTA \cite{he2020multi} & 70.8 & \textbf{94.9} & 85.2 & 92.7 & \textbf{92.6} & 71 \\
            \cline{2-8}
            & \textbf{DyGLIP} & \textbf{76.2} & 86.7 & \textbf{97.0} & \textbf{93.4} & 86.8 & \textbf{37} \\
        \hline
         Dataset2 & EGM \cite{chen2016equalized} & 47.9 & 67.2 & 70.6 & 79.8 & 63.3 & 1985 \\
            & RAC \cite{cai2014exploring} & 62.6 & 86.8 & 73.7 & 69.5 & 78.4 & 171 \\
            & ICLM \cite{lee2017online} & 67.7 & 88.3 & 76.6 & 83.3 & 70.9 & 123 \\
            & TRACTA \cite{he2020multi} & 83.7 & 93.4 & 85.9 & 95.5 & 95.4 & \textbf{60} \\
            \cline{2-8}
            & \textbf{DyGLIP} & \textbf{91.9} & \textbf{95.7} & \textbf{96.8} & \textbf{97.7} & \textbf{95.9} & {101} \\
        \hline
         Dataset3 & EGM \cite{chen2016equalized} & 18.6 & 27.0 & 64.7 & 82.1 & 53.5 & 525 \\
            & RAC \cite{cai2014exploring} & 5.6 & 9.2 & 55.3 & 47.5 & 66.2 & 666 \\
            & ICLM \cite{lee2017online} & 37.2 & 53.2 & 69.1 & 66.0 & 72.6 & 228 \\
            & TRACTA \cite{he2020multi} & 53.8 & 58.5 & 75.4 & 75.2 & 91.3 & 144 \\
            \cline{2-8}
            & \textbf{DyGLIP} & \textbf{89.4} & \textbf{92.7} & \textbf{96.5} & \textbf{98.2} & \textbf{93.7} & \textbf{122} \\
        \hline
         Dataset4 & EGM \cite{chen2016equalized} & 28.4 & 35.8 & 71.1 & 83.6 & 61.9 & 3111 \\
            & RAC \cite{cai2014exploring} & 34.0 & 53.9 & 63.0 & 52.2 & 79.4 & 329 \\
            & ICLM \cite{lee2017online} & 54.3 & 62.5 & 86.8 & 87.6 & 86.0 & 189 \\
            & TRACTA \cite{he2020multi} & 71.5 & 79.6 & 90.0 & 86.3 & \textbf{96.0} & \textbf{70} \\
            \cline{2-8}
            & \textbf{DyGLIP} & \textbf{84.7} & \textbf{92.5} & \textbf{96.6} & \textbf{91.3} & {92.9} & {100} \\
        \hline
    \end{tabular}
    \caption{Evaluation results on MCT dataset.}
    \vspace{-4mm}
    \label{tab:MCT_results}
\end{table*}
\begin{table}[!t]
\footnotesize
    \centering
    \resizebox{\linewidth}{!}{%
    \begin{tabular}{|c|c|c|c|}
    \hline
    \textbf{Subset} & \textbf{Method} & \textbf{MOTA} (\%) $\uparrow$ & \textbf{ID F1} (\%) $\uparrow$ \\ %
    \hline
    S02         & ELECTRICITY \cite{Qian_2020_CVPR_Workshops} & 53.7 & 53.8 \\
    \cline{2-4}
                 & \textbf{DyGLIP} & \textbf{90.9} & \textbf{64.9} \\ % & &  \\
    \hline
    S05         & ELECTRICITY \cite{Qian_2020_CVPR_Workshops} & 74.1 & 36.4 \\
    \cline{2-4}
                 & \textbf{DyGLIP} & \textbf{84.6} & \textbf{39.90} \\ % & &  \\
    \hline
    \end{tabular}}
    \caption{Results on CityFlow \cite{tang2019cityflow} validation set}
    \label{tab:AICity_resuls}
    \vspace{-4mm}
\end{table}
\textbf{Influence of Initial Node Embedding Features.}
We study the effects of original embedding features for each tracker/node used as inputs to the attention module in the proposed DyGLIP in Table \ref{tab:feature_reid}. We compare between using ReID features with the pre-trained model in \cite{zhou2019osnet} and features from detector, i.e., Mask-RCNN human detector \cite{he2017mask} on human MC-MOT. Note that all other experiments for human MC-MOT (except vehicle MC-MOT), we use these ReID features as inputs to our attention module.

\subsection{Comparison with State-of-the-arts MC-MOT}

To evaluate the proposed method, we first compare with other state-of-the-arts using three datasets PETS09 \cite{PETS09}, CAMPUS \cite{xu2016multi}, and EPFL \cite{EPFL} that contain videos with overlapping FOVs. Then, we also compare with other methods that do not require overlapping FOVs among different cameras on MCT \cite{chen2016equalized} dataset.
Finally, we perform evaluation on the validation set of CityFlow dataset \cite{tang2019cityflow} to compare with the winner of the 4\textsuperscript{th} AI City Challenge 2020 \cite{Naphade20AIC20}.
Although our method focuses on the online association task, our results are comparable with offline approaches. Besides, our DyGLIP method works well on both overlapping and non-overlapping fields of view by treating two types equally and solving the assignment task in a general way. 

\textbf{Overlapping FOVs dataset on Human Tracking}
We compare with other MC-MOT methods, including K-Shortest Path (KSP) \cite{berclaz2011multiple}, Hierarchical Composition of Tracklet (HCT) \cite{xu2016multi}, Brand-and-Price (B\&P) \cite{rosenhahn2012branch} and Spatio-Temporal Parsing (STP) \cite{xu2017cross} that require overlapping FOVs among different camera views.  
We also compare with TRACTA  \cite{he2020multi} on overlapping FOVs videos.
Tables \ref{tab:PETS09_results}, \ref{tab:epfl_results} and \ref{tab:campus_results} show the results on PETS09, CAMPUS, and EPFL datasets, respectively. DyGLIP outperforms all the other methods with up to 63 \% on certain metrics.

\textbf{Non-overlapping FOVs dataset on Human Tracking}
This experiment compares DyGLIP with other MC-MOT methods, including EGM \cite{chen2016equalized}, RAC \cite{cai2014exploring}, ICLM \cite{lee2017online} and TRACTA \cite{he2020multi}. These methods do not require overlapping FOVs between different camera views. DyGLIP significantly outperforms all the methods with a large margin (up to 35\%) in most metrics on MCT dataset as in Table \ref{tab:MCT_results}.

\textbf{Non-overlapping FOVs dataset on Car/Vehicles Tracking}
Table \ref{tab:AICity_resuls} shows the results on the AI City challenge validation set. We use the same ReID features as in ELECTRICITY \cite{Qian_2020_CVPR_Workshops}. Indeed, DyGLIP obtains much better results, higher on MOTA and ID F1 in S02 (37.2 \% and 11.1\%, respectively), in S05 (10.5 \% and 3.5\%, respectively), thanks to the dynamic graph formulation and the attention module.

\vspace{-2mm}
\section{Conclusion}
\vspace{-2mm}
This paper has re-formulated the MC-MOT problem with a dynamic graph model and treated the global tracklet ID association between multi-camera as link assignment in the proposed graph.
The robustness of the proposed dynamic graph is further improved with attention modules that capture structural and temporal variations across multi-camera and multiple time steps.
The experiments show significant performance improvements in both human and vehicle tracking datasets in multi-camera with overlapping and non-overlapping FOVs settings.

\section*{Acknowledgement}

We would like to thank Huy Bui, an AI engineer at VinAI Research, for helping with the baseline experiments.

{\small
\bibliographystyle{ieee_fullname}
\bibliography{egbib}
}

\end{document}